\documentclass[lettersize,journal]{IEEEtran}
\usepackage{amsmath,amsfonts}
\usepackage{algorithmic}
\usepackage{algorithm}
\usepackage{array}
\usepackage[caption=false,font=normalsize]{subfig}
\usepackage{textcomp}
\usepackage{stfloats}
\usepackage{url}
\usepackage{verbatim}
\usepackage{graphicx}
\usepackage{cite}
\usepackage{amssymb}
\usepackage{xcolor}
\usepackage{booktabs}
\usepackage{tabularx}
\usepackage[hidelinks]{hyperref}
\graphicspath{ {images/} }
\begin{document}

\title{FAM-DQ: A Dual-Quadrotor-Based Fully Actuated Aerial Manipulator for High-Torque Interaction}

\author{Xuwei Yang, Ruoyu Ren, Ziqian Guo}

\maketitle

\begin{abstract}
Aerial physical interaction requires aerial manipulation platforms to generate large interaction forces and torques while maintaining precise end-effector control. However, conventional underactuated aerial manipulators suffer from strong position-attitude coupling, whereas fully actuated platform designs often face structural complexity, limited payload capacity, and insufficient torque output. This paper presents FAM-DQ, a dual-quadrotor based fully actuated aerial manipulator designed for high-torque physical interaction tasks. By mounting two quadrotor propulsion modules at the ends of a central frame through passive joints, while using a gear-driven servo to regulate the pointing direction, FAM-DQ achieves decoupled $6$-DoF end-effector control with omnidirectional manipulation capability and enhanced torque output. Experiments including trajectory tracking, attitude tracking, static torque measurement, and screw driving validate the proposed design. FAM-DQ achieves a maximum torque of $1.019~\mathrm{N}\cdot\mathrm{m}$ with a total mass of $0.447~\mathrm{kg}$, corresponding to a torque-to-mass ratio of $2.28~\mathrm{N}\cdot\mathrm{m/kg}$.
\end{abstract}


\section{Introduction}
\IEEEPARstart{U}{ncrewed} aerial vehicles (UAVs) are evolving from non-contact sensing platforms to agents capable of active physical interaction. This paradigm, known as aerial physical interaction (APhI), enables complex industrial maintenance tasks, such as bolt tightening and valve operations\cite{ikedaWallContactOctorotor2017}. These tasks demand substantial torque to overcome environmental resistance, while maintaining precise, decoupled, and omnidirectional control of the end-effector pose. Consequently, balancing this high-torque output with precise end-effector control is a primary challenge for the practical deployment of aerial manipulation platforms.

A prevailing configuration in aerial manipulation is the underactuated aerial manipulator, characterized by a multi-degree-of-freedom (DoF) robotic arm integrated with a standard multirotor as a base platform. Benefiting from the maturity of both robotic arm and multirotor technologies, these platforms have been extensively applied to fundamental interaction tasks\cite{7989314,10547187,10722859,dingChatPMClassComposite2024}. Nevertheless, the physical interaction performance of such platforms is intrinsically constrained by the underactuated nature of the base platform. Specifically, the strong coupling between translational and rotational dynamics limits operational accuracy and compromises the base platform's capacity for disturbance rejection.

To overcome the coupling constraints of underactuated platforms, fully actuated and overactuated aerial manipulators have been widely investigated\cite{10610711,8629273,s19061305,8336503,malczykMultidirectionalInteractionForce2023,11247292,10902413,liFixedTimeControlNovel2024,parkODARAerialManipulation2018}. By using tilt-rotor or non-coplanar configurations, these platforms achieve full $6$-DoF controllability, enabling position-attitude decoupling and omnidirectional force generation during aerial manipulation. However, such designs typically introduce increased structural complexity and parasitic mass from additional actuators, which limit payload capacity and flight efficiency. Moreover, their limited geometric moment arms restrict interaction torque generation, making high-torque tasks such as valve turning prone to actuator saturation or stability degradation.

To circumvent the physical limitations of single-agent platforms and satisfy the demanding requirements of high-torque manipulations, researchers have investigated multi-agent collaborative or modular composite systems\cite{8460713,Nishio2024DesignCA,9720963,sugitoAerialManipulationUsing2022}. By mechanically linking multiple flight units, these platforms leverage distributed thrust and extended physical dimensions to amplify their manipulation capacity. Nevertheless, existing coupled platforms often exhibit severe kinematic and dynamic complexity. This complexity renders precise attitude control and trajectory tracking highly challenging, thereby limiting the practical deployment and reliability of these platforms.

To address these challenges, this paper proposes a dual-quadrotor-based fully actuated aerial manipulation platform, named \textbf{FAM-DQ} (A \textbf{D}ual-\textbf{Q}uadrotor-Based \textbf{F}ully \textbf{A}ctuated Aerial \textbf{M}anipulator for High-Torque Interaction). Unlike traditional aerial manipulators, FAM-DQ features two quadrotor propulsion modules mounted at the platform's ends via passive joints. This configuration achieves 6-DoF manipulation at the end-effector while utilizing the central frame's length to amplify the torque induced by the rotor thrusts.

The main contributions of this work are threefold.
\textit{(i)} A dual-quadrotor-based aerial manipulation platform is developed. The design utilizes the central frame as a physical lever to significantly amplify torque output, while enabling fully actuated end-effector control and omnidirectional manipulation capability.
\textit{(ii)} The dynamic model of the platform is derived, and a lightweight 6-DoF controller is designed to ensure accurate end-effector control.
\textit{(iii)} A series of experiments is conducted to validate the system. The results demonstrate the omnidirectional capabilities of the end-effector and the high torque output of the platform.

The remainder of this paper is organized as follows. Section~II presents the mechanical design of FAM-DQ; Section~III establishes the dynamic models of the entire system and propulsion modules; Section~IV describes the design of the controller; Section~V demonstrates the experimental validation results; and Section~VI concludes the paper.

\section{Mechanical Design}

\begin{figure}[t]
\centering
\includegraphics[width=0.45\textwidth]{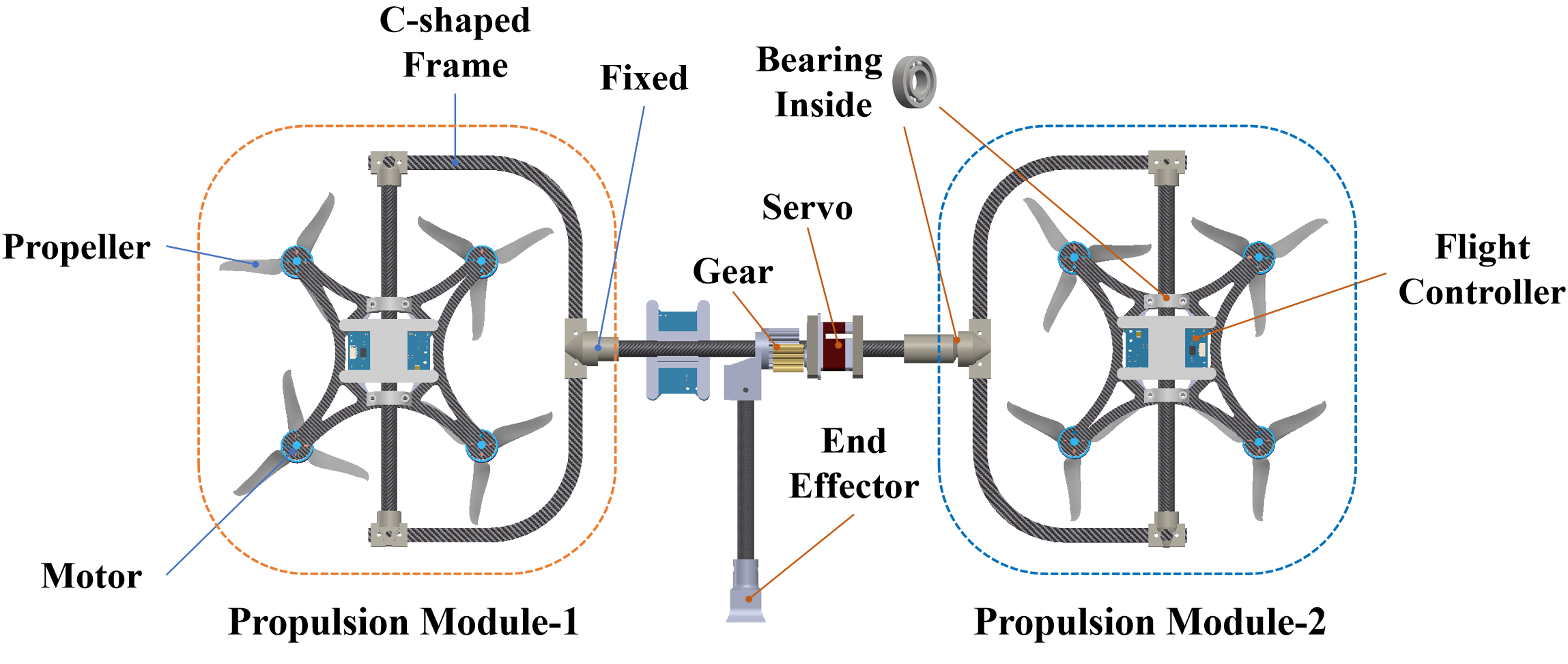}
\caption{Mechanical structure of FAM-DQ, including the central frame, propulsion modules, passive revolute joint, and end-effector.}
\label{fig:mechanical_design}
\end{figure}

FAM-DQ is designed to generate high interaction torque while maintaining decoupled end-effector control during aerial manipulation. The key mechanical concept is to combine lever-based torque amplification with a lightweight fully actuated configuration. As shown in Fig. \ref{fig:mechanical_design}, two quadrotor propulsion modules are arranged at both ends of a central frame. By using the central frame as a physical lever arm, the thrust generated by the propulsion modules can be converted into a large torque about the end-effector, which is particularly beneficial for high-torque tasks such as valve turning and screw driving. Compared with conventional non-coplanar fully actuated platforms, this configuration reduces internal force loss. It also avoids the parasitic mass introduced by actively tilting mechanisms.

The prototype consists of a central frame, an end-effector, and two propulsion modules. The central frame is built around a high-stiffness carbon-fiber tube, which provides both structural rigidity and an extended moment arm for torque generation. Driven by a gear-driven servo, the end-effector can rotate around the main rod to adjust its pointing direction over a wide range, enabling alignment with arbitrary task-relevant directions and satisfying the requirements of most aerial manipulation tasks.

To achieve controllable end-effector pointing while preserving the thrust-direction freedom of the propulsion modules, FAM-DQ adopts an asymmetric connection topology. C-shaped frames are mounted at both ends of the central frame to connect the propulsion modules. Propulsion module-1 is connected to its C-shaped frame through a passive joint, while this C-shaped frame is rigidly attached to the central frame. In contrast, the C-shaped frame associated with propulsion module-2 is connected to the central frame through a passive revolute joint. This topology releases the rigid attitude constraints of the propulsion modules, allowing their thrust vectors to be oriented for wrench generation. Meanwhile, propulsion module-1 and the gear-driven servo jointly regulate the end-effector pointing direction. The resultant end-effector force is generated by the vector sum of the two module thrusts, while the torque is produced by their moments about the end-effector, amplified by the large moment arm of the central frame. By coordinating the thrust magnitudes and directions of the two modules, FAM-DQ can generate the desired $6$-DoF wrench at the end-effector.

Each propulsion module consists of four 1204 $5000~\text{kV}$ brushless motors with 3015 propellers, an independent flight controller, electronic speed controllers, and an onboard battery. The two propulsion modules communicate with the central flight controller through a high-speed wireless link with a latency of approximately $10~\text{ms}$, enabling real-time transmission of thrust commands and module-level control information.

\section{Dynamic Model}
Propulsion module-2 incorporates two axial passive joints, enabling the generation of omnidirectional thrust vectors within the central frame, and propulsion module-1 is equipped with a single axial passive joint, which restricts the direction of the generated thrust to a two-dimensional plane within the central frame. The thrust produced by these two modules exerts significant torques on the system, particularly given the extended length of the central frame. Consequently, the formulation of the system dynamics is divided into two primary components. The translational and rotational dynamics of the entire system are derived first, followed by the individual rotational dynamics of each propulsion module.

\subsection{System Dynamics and Allocation}
\begin{figure}[t]
\centering
\includegraphics[width=0.45\textwidth]{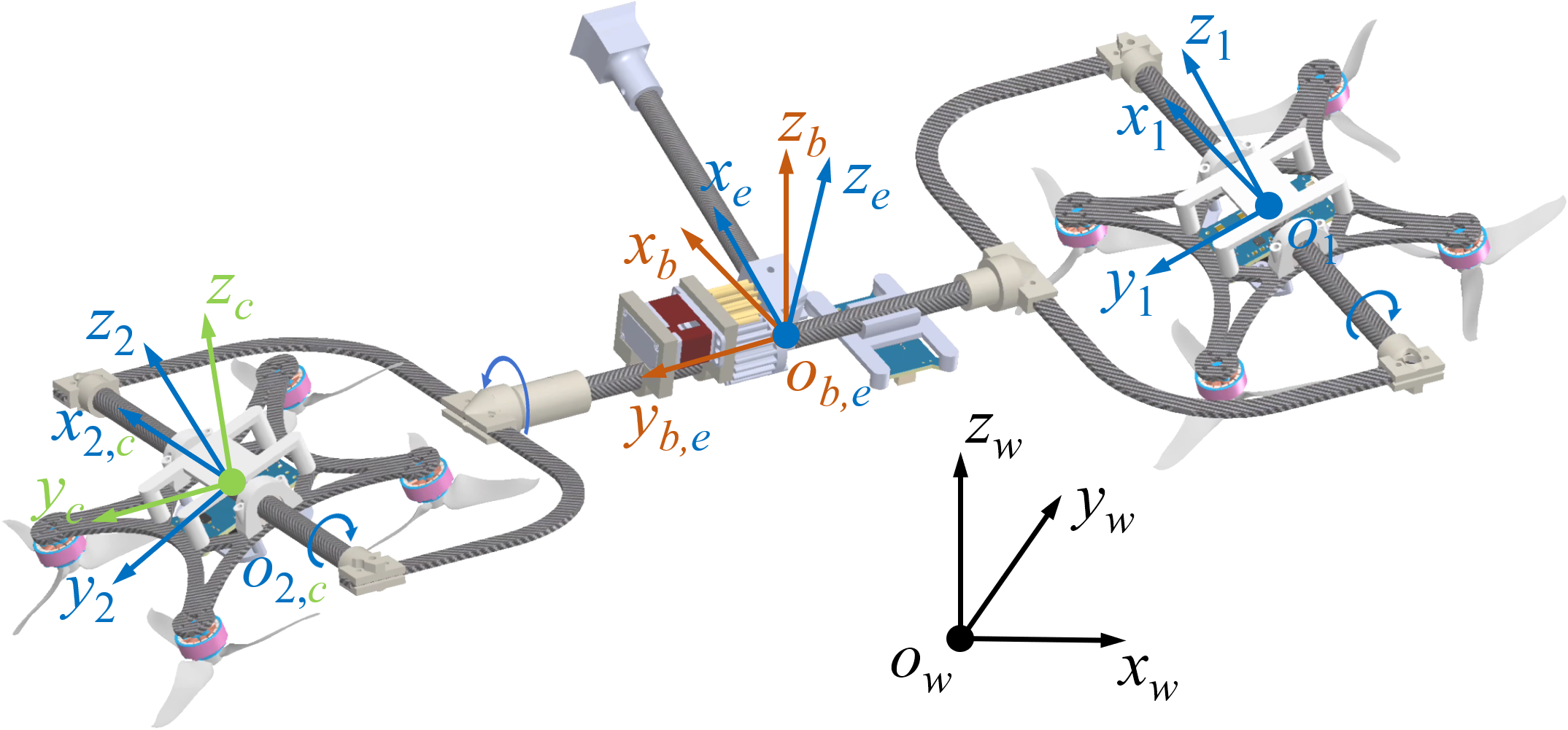}
\caption{Definitions of the coordinate frames. The frames $\mathcal{F}_w$, $\mathcal{F}_e$, and $\mathcal{F}_b$ represent the world, end-effector, and central frame body coordinate frames, respectively. The frames $\mathcal{F}_1$ and $\mathcal{F}_2$ denote the coordinate frames of two propulsion modules, while $\mathcal{F}_c$ indicates the coordinate frame of the C-shaped frame connected to propulsion module-2.}
\label{fig:coordinate}
\end{figure}

The coordinate frames are established as illustrated in Fig. \ref{fig:coordinate}. Within the central frame coordinate frame $\mathcal{F}_b$, the $\boldsymbol{y}_b$-axis aligns with the axial direction of the central frame, pointing toward the origin of propulsion module-2. The $\boldsymbol{z}_b$-axis is oriented upward, perpendicular to the C-shaped frame that is rigidly attached to the central frame. The $\boldsymbol{x}_b$-axis is then defined by the cross product of $\boldsymbol{y}_b$ and $\boldsymbol{z}_b$ in accordance with the right-hand rule.

Let $\boldsymbol{R}_b^a$ represent the rotation matrix of frame $b$ relative to frame $a$, and let $\boldsymbol{r}^a$ denote the coordinates of a vector $\boldsymbol{r}$ expressed in frame $a$. Based on the Newton-Euler equations, the system dynamics are formulated as
\begin{equation}
\label{sysdn}
\begin{aligned}
\ddot{\boldsymbol{p}} &= -g\boldsymbol{e}_3 + \frac{1}{M} \boldsymbol{R}_e^w \boldsymbol{f}^e \\
\dot{\boldsymbol{R}}_b^w &= \boldsymbol{R}_b^w [\boldsymbol{\omega}^b]_{\times} \\
\dot{\boldsymbol{\omega}}^b &= \boldsymbol{J}^{-1} \left( \boldsymbol{R}_e^b \boldsymbol{\tau}^e - \boldsymbol{\omega}^b \times \boldsymbol{J} \boldsymbol{\omega}^b \right)
\end{aligned}
\end{equation}
where $\boldsymbol{e}_3=[0,0,1]^T$ is the unit vector along the z-axis, $g$ denotes the gravitational acceleration, and $M$ represents the total mass of the system. The vector $\boldsymbol{p} \in \mathbb{R}^3$ indicates the position of the center of mass in $\mathcal{F}_w$. The vector $\boldsymbol{\omega}^b \in \mathbb{R}^3$ is the angular velocity of the central frame expressed in $\mathcal{F}_b$, and $\boldsymbol{J} \in \mathbb{R}^{3 \times 3}$ is the inertia tensor of the system with respect to $\mathcal{F}_b$. Furthermore, $\boldsymbol{f}^e \in \mathbb{R}^3$ and $\boldsymbol{\tau}^e \in \mathbb{R}^3$ denote the thrust and torque exerted by the propulsion modules in $\mathcal{F}_e$. The torque term incorporates the moments induced by the propulsion modules thrusts and torques.

To establish a dynamic model suitable for analysis and controller design, several simplifying assumptions are introduced. First, the influence of the motion of the end-effector on the overall center of mass and inertia tensor is neglected. Second, since the propulsion modules are located far from the center of mass of the system, they are approximated as point masses. Therefore, the slight variations in the overall inertia tensor induced by the rotation of the propulsion modules about the passive joints are omitted. The total thrust $\boldsymbol{f}^e$ generated by the propulsion modules in \eqref{sysdn} is expressed in the end-effector frame $\mathcal{F}_e$. This representation is adopted to facilitate the analysis of the thrust capabilities of the system in $\mathcal{F}_e$ rather than in the world frame $\mathcal{F}_w$ in subsequent sections.

Prior to the derivation of the thrust, the attitudes and control inputs of the propulsion modules are defined. As illustrated in Fig. \ref{fig:coordinate}, the frames of the propulsion modules are denoted by $\mathcal{F}_1$ and $\mathcal{F}_2$. Based on the specific joint designs of the two modules, the attitudes of the propulsion modules are parameterized relative to $\mathcal{F}_e$. Letting $\boldsymbol{R}_x(\cdot)$ and $\boldsymbol{R}_y(\cdot)$ denote the rotation matrices about the $x$- and $y$-axes, respectively, the relative attitude is given by
\begin{equation}
\boldsymbol{R}_i^e = \boldsymbol{R}_y(\theta_i) \boldsymbol{R}_x(\phi_i), \quad i=1,2.
\end{equation}
The control input of each propulsion module comprises the thrust along the $\boldsymbol{z}_i$-axis and torques in the propulsion module body frame, denoted by $\boldsymbol{u}_i = [f_i, \boldsymbol{\tau}_i^T]^T$ for $i=1,2$. The torques about the $\boldsymbol{x}_i$- and $\boldsymbol{y}_i$-axes are induced by the rotor thrusts and are proportional to the arm length, whereas the torque about the $\boldsymbol{z}_i$-axis is generated by the aerodynamic reaction torque of the rotors. 

Returning to \eqref{sysdn}, the total thrust generated by the propulsion modules is expressed as
\begin{equation}
\label{fm}
\boldsymbol{f}^e = \boldsymbol{R}_1^e \boldsymbol{e}_3 f_1 + \boldsymbol{R}_2^e \boldsymbol{e}_3 f_2.
\end{equation}

The torque generated by the propulsion modules is formulated as follows. Since two propulsion modules are symmetrically mounted on opposite sides of the central frame at a distance $L$ from the origin $O_b$, the total torque $\boldsymbol{\tau}^b$ consists of two primary components. These components include the torque induced by thrust and torque of the propulsion modules. This relationship is expressed as
\begin{equation}
\label{taub}
\begin{aligned}
\boldsymbol{\tau}^{b} &= L \boldsymbol{e}_2 \times \left( -\boldsymbol{R}_1^b \boldsymbol{e}_3 f_1 + \boldsymbol{R}_2^b \boldsymbol{e}_3 f_2 \right)\\
&\quad + \boldsymbol{P} \boldsymbol{R}_1^b \boldsymbol{\tau}_1 + \boldsymbol{R}_y(\theta_2-\theta_1) \boldsymbol{P} \boldsymbol{R}_x(\phi_2) \boldsymbol{\tau}_2
\end{aligned}
\end{equation}
Here, $\boldsymbol{e}_2=[0,1,0]^T$ denotes the unit vector along the $y$-axis. The cross-product terms represent the torque induced by the thrust $f_i$ of each module. The opposite signs result from the placement of the two modules in opposing directions along the $\boldsymbol{y}_b$-axis. The subsequent terms denote the torque transferred from the propulsion modules to the central frame, where $\boldsymbol{P} = \text{diag}(0, 0, 1)$ serves as the torque transmission matrix. By modeling the two modules as point masses at the ends of the central frame, the matrix $\boldsymbol{P}$ extracts the torque components along the $\boldsymbol{z}_b$ and $\boldsymbol{z}_c$-axes. Although the theoretical moment of inertia of an equivalent point mass about the $\boldsymbol{y}_b$-axis is zero, the actual inertia of propulsion module-1 about this axis remains non-negligible. Consequently, the torque along the $\boldsymbol{y}_b$-axis is not extracted directly through $\boldsymbol{P}$, instead, it is quantified via \eqref{dot_omega_y}. Furthermore, the expression $\boldsymbol{P} \boldsymbol{R}_x(\phi_2) \boldsymbol{\tau}_2$ isolates the component of $\boldsymbol{\tau}_2$ along $\boldsymbol{z}_c$-axis. This component is subsequently transformed into $\mathcal{F}_b$ using the rotation matrix $\boldsymbol{R}_y(\theta_2-\theta_1)$. The total system input within the frame $\mathcal{F}_e$ is decoupled by defining the attitude of the central frame relative to the end-effector as $\boldsymbol{R}_b^e=\boldsymbol{R}_y(\theta_1)$ and applying the coordinate transformation $\boldsymbol{\tau}^e = \boldsymbol{R}_b^e \boldsymbol{\tau}^b$. By combining Equations \eqref{fm} and \eqref{taub}, the expanded scalar components of the system input are expressed as follows:
\begin{equation}
\label{fmtaum}
\left\{
\begin{aligned}
f_x^e &= f_1 \cos\phi_1 \sin\theta_1 + f_2 \cos\phi_2 \sin\theta_2 \\
f_y^e &= -f_1 \sin\phi_1 - f_2 \sin\phi_2 \\
f_z^e &= f_1 \cos\phi_1 \cos\theta_1 + f_2 \cos\phi_2 \cos\theta_2 \\
\tau_x^e &= \tau_{1z} \cos\phi_1 \sin\theta_1 + \tau_{2z} \cos\phi_2 \sin\theta_2 \\
&\quad + \tau_{1y} \sin\phi_1 \sin\theta_1 + \tau_{2y} \sin \phi_2 \sin \theta_2 \\
&\quad - Lf_1 \cos \phi_1 \cos \theta_1 + Lf_2 \cos \phi_2 \cos \theta_2 \\
\tau_z^e &= \tau_{1z} \cos \phi_1 \cos \theta_1 +\tau_{2z} \cos \phi_2 \cos \theta_2 \\
&\quad + \tau_{1y} \cos \theta_1 \sin \phi_1 +\tau_{2y} \cos \theta_2 \sin \phi_2 \\
&\quad + Lf_1 \cos \phi_1 \sin \theta_1 - Lf_2 \cos \phi_2 \sin \theta_2
\end{aligned}
\right.
\end{equation}
Since the end-effector is connected to the central frame via a servo, it is able to remain unaffected during the frame's rotation around $\boldsymbol{y}_b$. Consequently, this degree of freedom does not require active control but follows the rotation of propulsion module-1. Therefore, $\tau_y^e$ is not included in the above equation.

\subsection{Propulsion Modules Dynamics}

The dynamic model of the system was established in the previous subsection. To generate the required thrust and torque, it is necessary to control the attitude and thrust of the propulsion modules. Therefore, this subsection establishes the dynamic model of the propulsion modules. The translational dynamics of the propulsion modules follow the same as the system translational dynamics in previous subsection. Consequently, this section only establishes the independent rotational dynamics models of the two propulsion modules. 

The roll of propulsion module-2 is provided by two coaxial bearings mounted on a C-shaped frame, and the pitch is provided by a bearing on the central frame along $\boldsymbol{y}_b$-axis. When propulsion module-2 rotates around $\boldsymbol{y}_b$-axis, it causes the C-shaped frame to rotate simultaneously. Since the mass and the moment of inertia of the C-shaped frame are negligible compared to those of the module, this coupling effect is treated as a minor torque disturbance along $\boldsymbol{y}_b$-axis. Unlike the propulsion module-2, the rotation of propulsion module-1 around $\boldsymbol{y}_b$-axis is coupled with the rotation of the central frame. This motion generates an internal torque $\tau_y^e$ about the $\boldsymbol{y}_b$-axis. Additionally, the rotationals degree of freedom of both propulsion modules about the $\boldsymbol{z}_b$ or $\boldsymbol{z}_c$-axis are restricted, which introduces constraint torques. Although these constraint torques contribute to the dynamics of the system, as shown in Equation \eqref{taub}, the magnitude of these torques is much smaller than the torques generated by the thrust of two modules. Consequently, the system controller does not utilize the torques of the propulsion modules to control the attitude of the central frame, these constraint torques can be omitted from the dynamics model of the propulsion modules. Taking these conditions into consideration, the rotational dynamics of propulsion module-$i$ ($i=1,2$) are formulated as follows:
\begin{equation}
\label{quadrotordy}
\begin{aligned}
\dot{\boldsymbol{R}}_i &= \boldsymbol{R}_i [\boldsymbol{\omega}_i]_\times \\
\dot{\boldsymbol{\omega}}_i &= \boldsymbol{J}_i^{-1} \left( \boldsymbol{\tau}_i  + \tau_{d_i}\boldsymbol{R}_i^T \boldsymbol{e}_2 - \boldsymbol{\omega}_i \times \boldsymbol{J}_i \boldsymbol{\omega}_i \right)
\end{aligned}
\end{equation}
where $\boldsymbol{R}_i \in \mathbb{R}^{3 \times 3}$ denotes the attitude of the propulsion module-i frame $\mathcal{F}_i$ relative to $\mathcal{F}_w$. $\boldsymbol{\omega}_i \in \mathbb{R}^3$ represents the angular velocity expressed in $\mathcal{F}_i$, and $\boldsymbol{J}_i \in \mathbb{R}^{3 \times 3}$ is the inertia matrix expressed in $\mathcal{F}_i$. $\boldsymbol{\tau}_i \in \mathbb{R}^3$ represents the torque produced by the thrust and the drag of the rotors. The scalar $\tau_{d_i}$ represents the external disturbance torque applied to the respective propulsion module, with $\tau_{d_1} = \tau_y^e$.

Let $\boldsymbol{\varpi}_i \in \mathbb{R}^4$ denote the vector of squared motor speeds. Based on the thrust and torque model of a quadrotor, the relationship between the control input $\boldsymbol{u}_i$ for the propulsion module-$i$ and $\boldsymbol{\varpi}_i$ is formulated as
\begin{equation}
\label{ui}
\boldsymbol{u}_i = \boldsymbol{T} \boldsymbol{\varpi}_i
\end{equation}
where $\boldsymbol{T} \in \mathbb{R}^{4 \times 4}$ represents the rotor effectiveness matrix. Since propulsion modules have same rotor configurations, thrust coefficients, and torque coefficients, they share the same effectiveness matrix.

The rotational degree of freedom for propulsion module-1 around $\boldsymbol{y}_b$-axis is rigidly coupled to the central frame. Consequently, $\tau_y^e$ functions as both an external load on the propulsion module-1 and driving torque to rotate the central frame around $\boldsymbol{y}_b$. To facilitate dynamic decoupling and feedforward compensation in the subsequent controller design, it is necessary to establish the relevant kinematic relationships. The absolute angular velocity of propulsion module-1 expressed in the body frame $\mathcal{F}_b$, denoted by $\boldsymbol{\omega}_1^b$, is the sum of the absolute angular velocity of the central frame $\boldsymbol{\omega}^b$ and the relative angular velocity. This relationship yields $\boldsymbol{\omega}_1^b = \boldsymbol{\omega}^b + \dot{\phi}_1 \boldsymbol{e}_1$. Taking the time derivative of this equation and extracting the component along $\boldsymbol{y}_b$-axis provides the kinematic constraint for the angular accelerations:
\begin{equation}
\label{dot_omega_y}
\dot{\omega}_{y}^b = \dot{\omega}_{1_y}^b - \omega_{z}^b \dot{\phi}_1
\end{equation}

In this section, we have developed the system dynamic model and obtained the mapping between the total thrust/torque and the thrust, torque, and attitude of the propulsion modules. Additionally, individual propulsion module models were established to facilitate the generation of the required thrust and torque.

\section{Controller Design}
\begin{figure*}[t]
\centering
\includegraphics[width=0.9\textwidth]{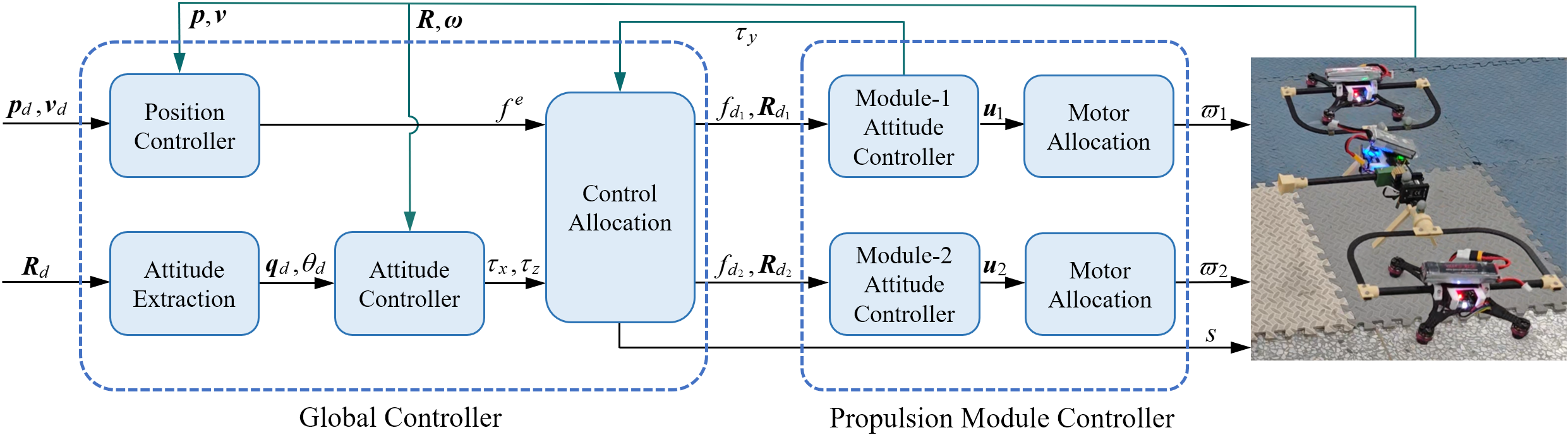}
\caption{Architecture of the proposed control system.}
\label{fig:controllerdiagram}
\end{figure*}

This section presents a cascaded control architecture based on the previously established dynamics model. As illustrated in Fig. \ref{fig:controllerdiagram}, the controller comprises two primary components: a global controller for the system and local controllers for the propulsion modules. The global controller computes the desired total thrust and torque, then a control allocation module maps the thrust and torque into desired attitude and thrust commands for the individual propulsion modules. Finally, local geometric controllers within the propulsion modules track these commands.

\subsection{Global Controller}
Due to the fully actuated nature of the system, the global controller consists of decoupled position and attitude controllers. These controllers utilize the desired reference states and feedback states to generate the required thrust and torque. Based on the established dynamics model, the two propulsion modules function as actuators that provide omnidirectional thrust, and therefore the control allocation module processes the desired total thrust and torque to compute the necessary attitude and thrust setpoints for each propulsion module.

\textbf{Position Controller}: To compute the desired thrust, the position controller receives the desired position $\boldsymbol{p}_d$ and velocity $\boldsymbol{v}_d$ from a high-level planner, along with the current feedback states. It employs a cascaded PID structure formulated as
\begin{equation}
u = k_p ( \lambda_d - \lambda ) + k_d ( \dot{\lambda}_d - \dot{\lambda} ) + k_i \int ( \lambda_d - \lambda ) dt
\end{equation}
where $\lambda \triangleq \left\{ p_x^w,p_y^w,p_z^w,v_x^w,v_y^w,v_z^w \right\}$ represents the relevant state variables, and $u$ denotes the corresponding control output. Following a coordinate transformation, the output of the velocity loop serves as the force vector $\boldsymbol{f}^e$ in \eqref{fmtaum}. The parameters $k_p$, $k_d$, and $k_i$ denote the controller gains. Additionally, a gravity compensation term is incorporated into the final output of the velocity loop.

\textbf{Attitude Extraction}: The orientation control of the end-effector is decomposed into controlling the pointing of the main rod, $\boldsymbol{q}$, and the servo-actuated pointing $\theta_s$. Therefore, $\boldsymbol{q}_d$ and $\theta_d$ need to be calculated from the desired end-effector attitude.

\textbf{Attitude Controller}: The attitude controller processes the desired attitude from the upper-level planner and the current state feedback to generate the desired two-axis torques and servo control signal. The servo angle command $s$ illustrated in Fig. \ref{fig:controllerdiagram} is directly determined by the error between the desired and current pitch angles of the central frame. For the pointing of main rod, let $\boldsymbol{q}$ and $\boldsymbol{q}_d$ denote the current and desired unit pointing vectors in $\mathcal{F}_w$, respectively. Since the bandwidth of the entire system is significantly lower than the bandwidth of the individual propulsion modules, the high-order nonlinear terms in \eqref{sysdn} are negligible. This justifies a cascade control scheme for the pointing of main rod. The desired angular velocity is defined as
\begin{equation}
\boldsymbol{\omega}_d^b = k_q\boldsymbol{R}_w^b ( \boldsymbol{q} \times \boldsymbol{q}_d )
\end{equation}
where $k_q$ is a gain that regulates the response speed. It can be noticed that the desired angular velocity is orthogonal to $\boldsymbol{q}$, therefore based on the dynamic model without nonlinear terms, a standard PID controller could be employed to compute the resulting torque vector $\tau_x^e$ and $\tau_z^e$, which is also orthogonal to $\boldsymbol{q}$. $\tau_y^e$ is coupled with and driven by propulsion module-1, will be given in the next subsection.

\textbf{Control Allocation}: During the control allocation, the desired thrust $\boldsymbol{f}^e$ and the torques $\tau_x^e, \tau_z^e$ are derived from the dynamic model in \eqref{fmtaum}. The body torques $\boldsymbol{\tau}_i$ of the propulsion modules are neglected at this stage, as the magnitudes of these torques are substantially smaller than the magnitudes of the primary torques induced by thrusts. After this simplification, \eqref{fmtaum} contains six variables corresponding to the thrust magnitudes and the angles of propulsion modules. It can be noticed that desired thrust along the $\boldsymbol{y}_b$-axis can be provided by either of the propulsion modules without causing coupling effects on other degrees of freedom. To resolve this over-actuation problem and to balance the energy consumption between the propulsion modules, the propulsion modules are constrained to produce equal thrust components along the $\boldsymbol{y}_b$-axis, which is formulated as
\begin{equation}
\label{f1f2}
f_1 \sin \phi_1 = f_2 \sin \phi_2
\end{equation}

Subject to the constraint in \eqref{f1f2}, the required thrust magnitudes $f_1, f_2$ and the desired attitude angles of propulsion modules can be obtained analytically. However, the solution for the angles $\theta_1, \phi_1, \theta_2, \phi_2$ is not unique, as an identical thrust vector can be generated by different tilt configurations. Under non-singular conditions, each propulsion module possesses two sets of solutions for $\phi_i$ and $\theta_i$. For instance, the configurations $f_1 = 1, \phi_1 = \pi/3, \theta_1 = \pi/4$ and $f_1 = 1, \phi_1 = 2\pi/3, \theta_1 = -3\pi/4$ yield the same thrust while representing distinct tilt configurations. Let $\mathcal{R}_{d_i} = \{\boldsymbol{R}_{d_i1}, \boldsymbol{R}_{d_i2}\}$ denote the set of feasible desired attitude matrices. To prevent aggressive maneuvers, the solution that is closest to the current attitude is selected as the final desired attitude, which is given by
\begin{equation}
\boldsymbol{R}_{d_i} = \arg\max_{\boldsymbol{R}_{d_ij} \in \mathcal{R}_{d_i}} \text{tr}(\boldsymbol{R}_i^T \boldsymbol{R}_{d_ij})
\end{equation}

\subsection{Propulsion Module Controller}
\textbf{Module Attitude Controller}: The mass and the moment of inertia of the attached C-shaped frame are negligible compared with those of propulsion module-2. Consequently, the disturbance torque $\boldsymbol{\tau}_{d_2}$ in \eqref{quadrotordy} is sufficiently small and can be safely neglected, relying on the inherent robustness of the controller. The controller of propulsion module-1 utilizes a similar structure but necessitates an additional torque $\tau_{d_1}$ along the $\boldsymbol{y}_b$-axis.

The desired attitude $\boldsymbol{R}_{d_i}$ is given in previous subsection, desired angular velocity $\boldsymbol{\omega}_{d_i} = [\boldsymbol{R}_{d_i}^T \dot{\boldsymbol{R}}_{d_i}]^\vee$, where $[\cdot]^\vee$ represents the vee operator from $\mathfrak{so}(3)$ to $\mathbb{R}^3$. Following the methodology presented in\cite{leeGeometricTrackingControl2010}, the attitude error $\boldsymbol{e}_{R_i} \in \mathbb{R}^3$ is defined as
\begin{equation}
\boldsymbol{e}_{R_i} = \frac{1}{2} \left( \boldsymbol{R}_i^T \boldsymbol{R}_{d_i} - \boldsymbol{R}_{d_i}^T \boldsymbol{R}_i \right)^\vee
\end{equation}
and the angular velocity error $\boldsymbol{e}_{\omega_i} \in \mathbb{R}^3$ is given by
\begin{equation}
\boldsymbol{e}_{\omega_i} = \boldsymbol{R}_i^T \boldsymbol{R}_{d_i} \boldsymbol{\omega}_{d_i} - \boldsymbol{\omega}_i
\end{equation}

Consequently, the torque input for propulsion module-$i$ is formulated as
\begin{equation}
\label{taui}
\boldsymbol{\tau}_i = \left( \boldsymbol{I} - \boldsymbol{z}_{c_i}^i {\boldsymbol{z}_{c_i}^{iT}} \right) \left(
\boldsymbol{k}_R \boldsymbol{e}_{R_i} + \boldsymbol{k}_\omega \boldsymbol{e}_{\omega_i} + \boldsymbol{\omega}_i \times \boldsymbol{J}_i \boldsymbol{\omega}_i \right)
\end{equation}
For a unified representation, the vector $\boldsymbol{z}_{c_i}^i$ denotes the $\boldsymbol{z}$-axis of the C-shaped frame connected to propulsion module-$i$, expressed in the body frame of each propulsion module. Specifically, $\boldsymbol{z}_{c_1}$ aligns with $\boldsymbol{z}_b$ for propulsion module-1, whereas $\boldsymbol{z}_{c_2}$ corresponds to $\boldsymbol{z}_c$, as depicted in Fig. \ref{fig:coordinate}. The matrix $\boldsymbol{I} - \boldsymbol{z}_{c_i}^i {\boldsymbol{z}_{c_i}^{iT}}$ serves as a projection operator. Under conditions without aggressive maneuvers, this operator effectively eliminates the torque component along the $\boldsymbol{z}_{c_i}$ axis while preserving the remaining components. This avoids torque generation between propulsion modules and the central frame, which would otherwise degrade the dynamics of the system. The matrices $\boldsymbol{k}_R$ and $\boldsymbol{k}_\omega$ represent positive definite diagonal gain matrices\cite{rajappaModelingControlDesign2015}.

Furthermore, since propulsion module-1 is rigidly coupled to the central frame along the $\boldsymbol{y}_b$-axis, it must generate the torque $\tau_{d_1}$ to drive the central frame. Based on the angular acceleration constraint in \eqref{dot_omega_y} and the system dynamics in \eqref{sysdn}, the required driving torque $\tau_{d_1}$ is computed and superimposed on the torque of propulsion module-1 in \eqref{taui}.

\textbf{Motor Allocation}: Given the control inputs $\boldsymbol{u}_i$ for each propulsion module, the desired squared speeds of the rotors, denoted by $\boldsymbol{\varpi}_i$, are calculated utilizing the inverse rotor effectiveness matrix $\boldsymbol{T}^{-1}$.

\section{Experiments}
This section presents a series of experiments to validate the fully-actuation capabilities and manipulation performance of the FAM-DQ. These experiments involve the decoupled tracking of spatial trajectories and attitudes, alongside practical manipulation tasks. All experiments are conducted in an indoor environment, where a motion capture system provides real-time position feedback. A video demonstrating the experiments is provided as supplementary material.

\subsection{Position Tracking}
\begin{figure}[t]
\centering
\subfloat[]{\includegraphics[width=0.48\linewidth]{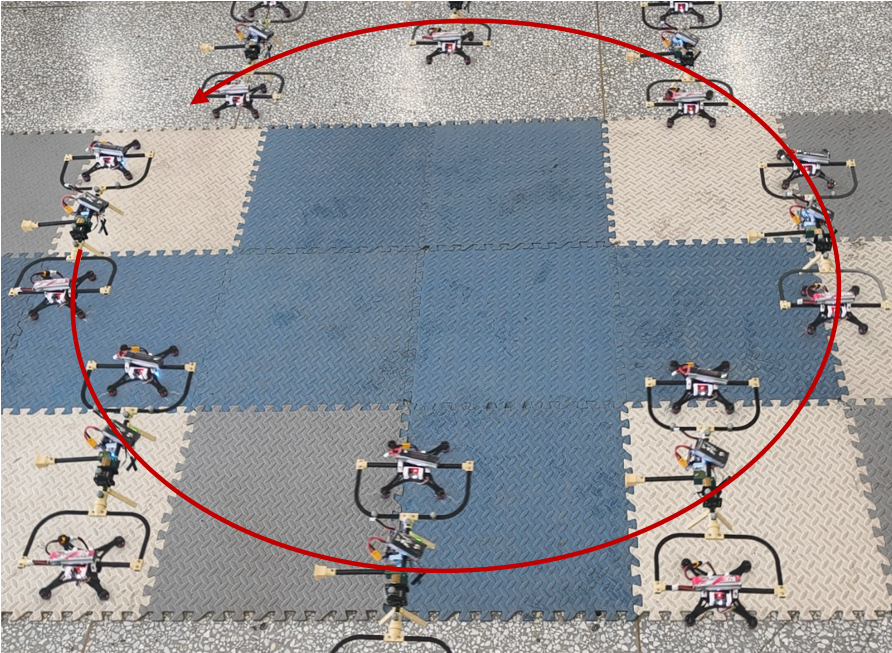}\label{fig:exp1_a}} \hfill
\subfloat[]{\includegraphics[width=0.48\linewidth]{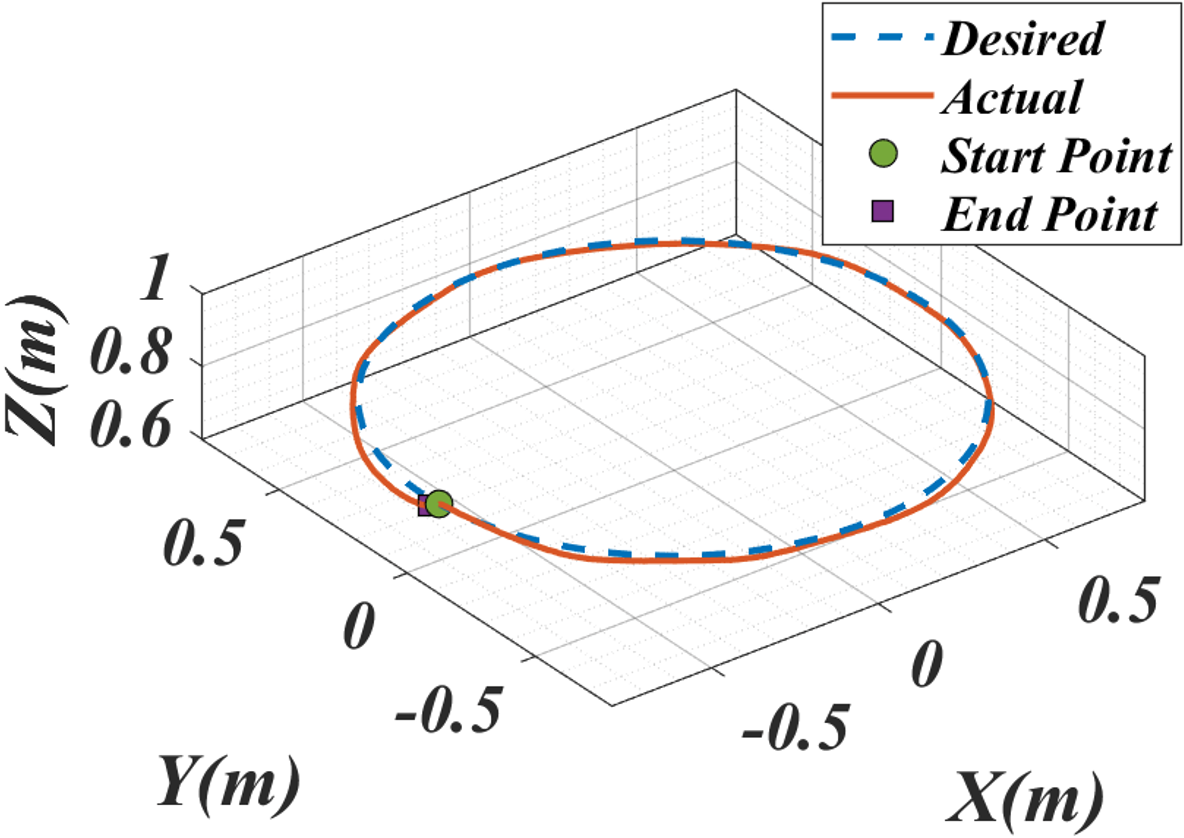}\label{fig:exp1_b}} \\[4pt]
\subfloat[]{\includegraphics[width=0.98\linewidth]{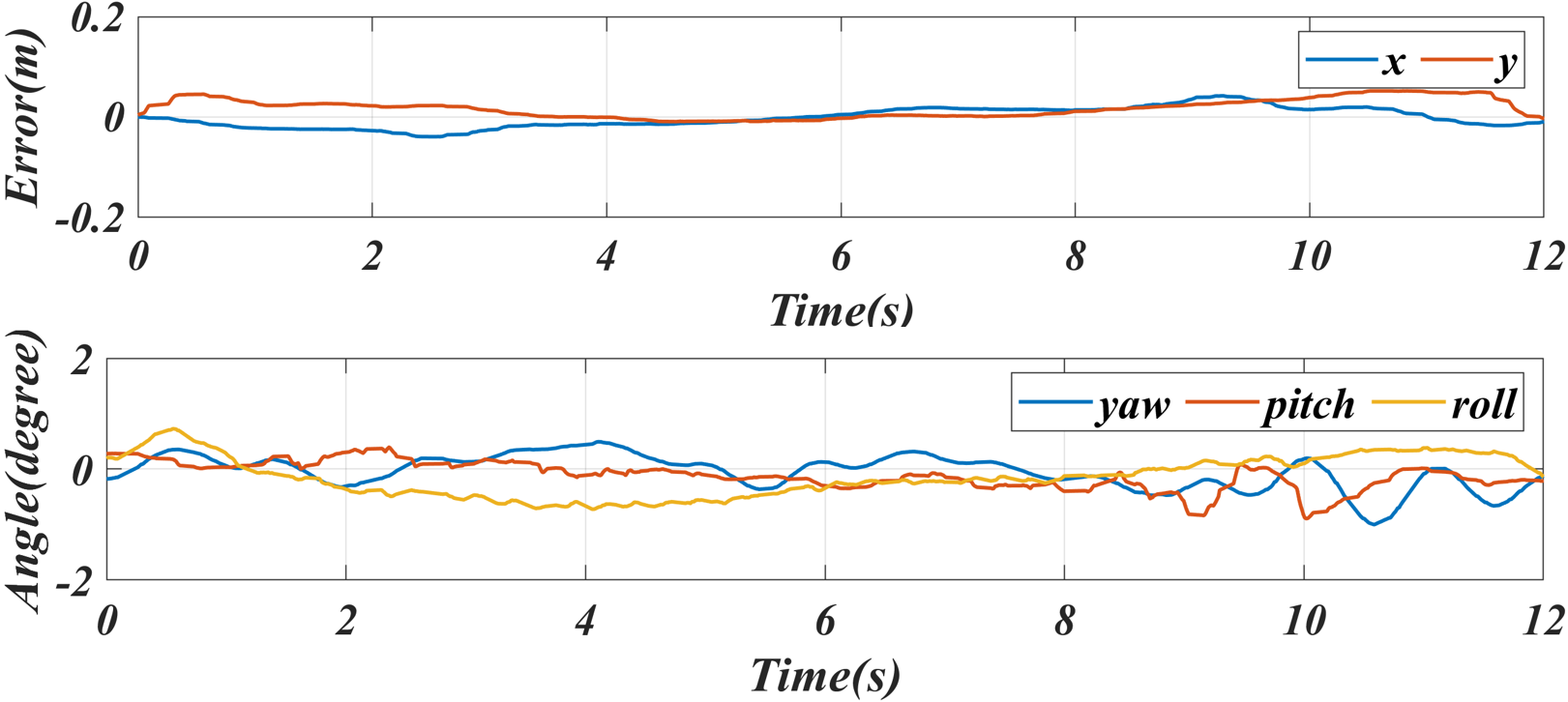}\label{fig:exp1_c}}
\caption{Circular trajectory tracking results of FAM-DQ while maintaining a constant attitude of the end-effector. (a) Experimental snapshot. (b) Comparison between the desired and actual trajectories. (c) Position errors and attitude angles during the tracking process.}
\label{fig:exp1}
\end{figure}

A circular trajectory tracking experiment is designed to validate the position-attitude decoupling capability of FAM-DQ, which is commanded to track a circular path with a radius of $0.75$ m in the $XY$-plane while maintaining a constant zero-attitude of the end-effector. The flight process is visualized in Fig. \ref{fig:exp1}\subref{fig:exp1_a}. As shown in Fig. \ref{fig:exp1}\subref{fig:exp1_b}, the actual trajectory closely follows the desired reference. Quantitative results in Fig. \ref{fig:exp1}\subref{fig:exp1_c} indicate a high tracking precision, with position errors consistently remaining within $\pm 0.05$ m. The attitude angles in Fig. \ref{fig:exp1}\subref{fig:exp1_c} demonstrate that despite the continuous translational motion, the attitude angles exhibit minimal fluctuations, bounded within $\pm 1.5^\circ$. These results demonstrate that FAM-DQ is capable of maintaining precise position control while keeping attitude unaffected, validating the effectiveness of the proposed actuation and allocation strategy.

\subsection{Attitude Tracking}

\begin{figure}[t]
    \centering
    \subfloat[]{\includegraphics[width=0.98\linewidth]{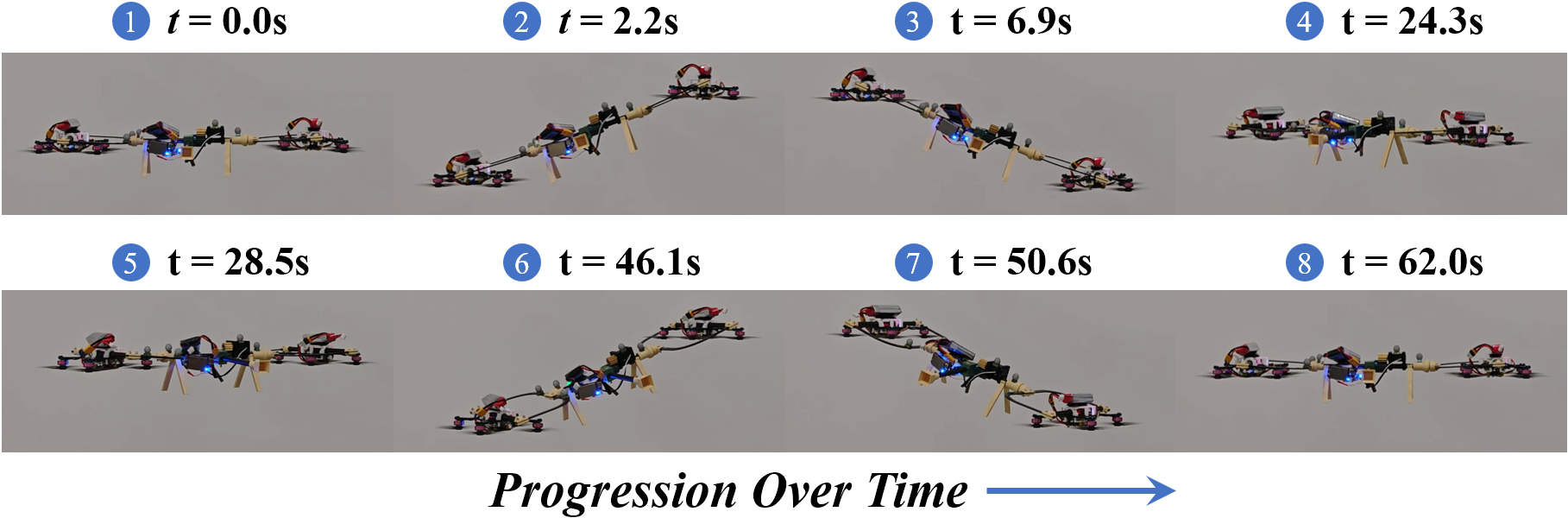}\label{fig:exp2_a}} \\[4pt]
    \subfloat[]{\includegraphics[width=0.98\linewidth]{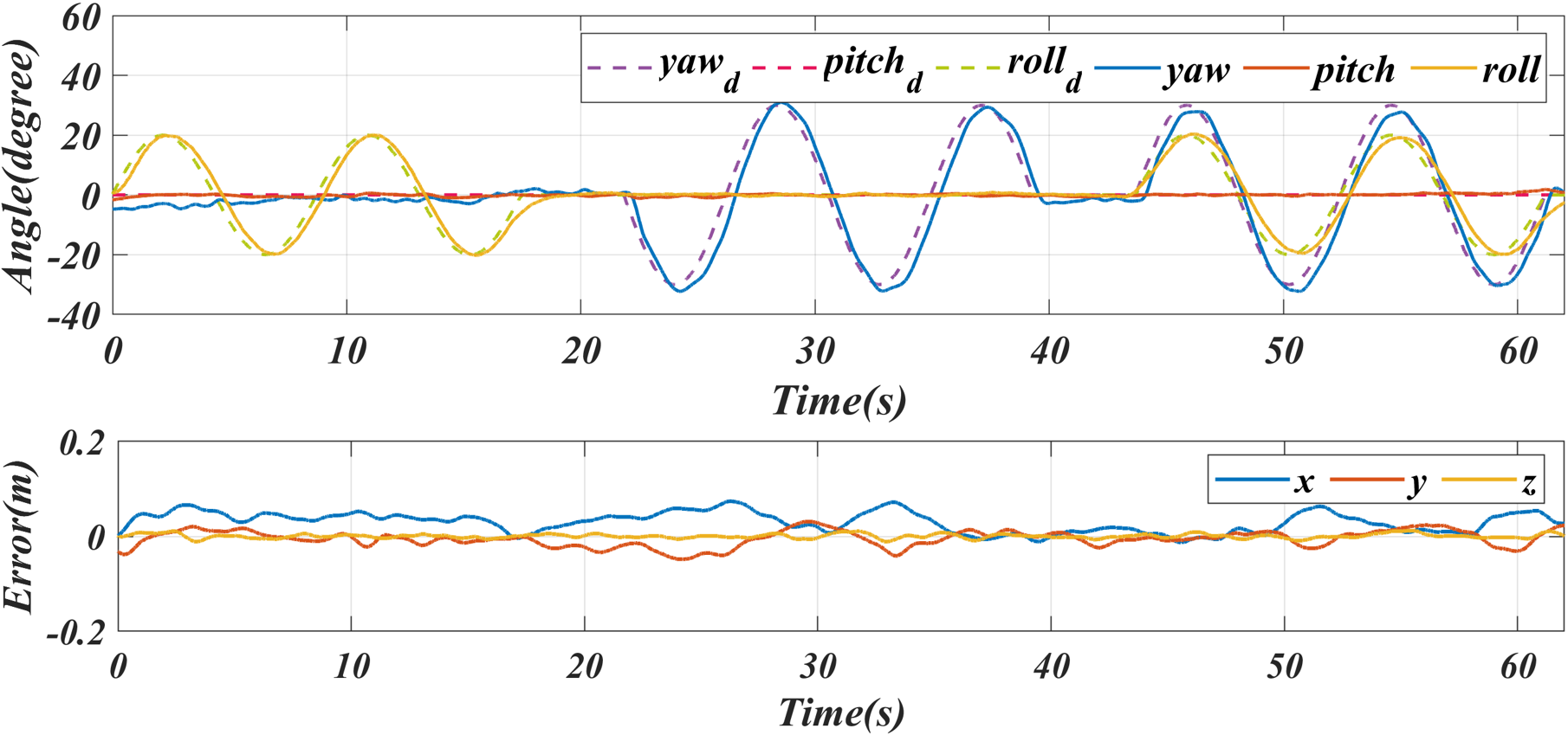}\label{fig:exp2_b}}
    \caption{Hovering attitude tracking experiment to verify the position-attitude decoupling capability of FAM-DQ. (a) Time-lapse snapshots of the FAM-DQ tracking time-varying orientations at a fixed hovering point. (b) Tracking curves for sinusoidal attitude commands and corresponding position errors during tracking.}
    \label{fig:exp2}
\end{figure}

Complementary to the first experiment, a hovering attitude tracking experiment is conducted to verify the position-attitude decoupling capability. FAM-DQ is commanded to track time-varying orientations while hovering at a fixed point, as shown in Fig. \ref{fig:exp2}\subref{fig:exp2_a}. As detailed in Fig. \ref{fig:exp2}\subref{fig:exp2_b}, the experimental demonstration consists of three consecutive phases: a pure sinusoidal roll with an amplitude of $20^\circ$, a pure sinusoidal yaw with an amplitude of $30^\circ$, and a challenging coupled roll-yaw maneuver. The responses demonstrate that FAM-DQ reliably reaches the desired attitudes. A small phase lag, mainly caused by actuator dynamics and signal latency, can be observed, but the system remains stable throughout the maneuver. Despite attitude variations of up to $30^\circ$, the translational position is well-maintained, as shown in Fig. \ref{fig:exp2}\subref{fig:exp2_b}. Position errors are bounded within $\pm 0.08$ m ($x$), $\pm 0.05$ m ($y$), and a notably small $\pm 0.02$ m ($z$). This precise position holding during aggressive attitude maneuvers validates FAM-DQ's independent 6-DoF control capability.

\subsection{Static Torque Output}
\begin{figure}[t]
\centering
\includegraphics[width=0.45\textwidth]{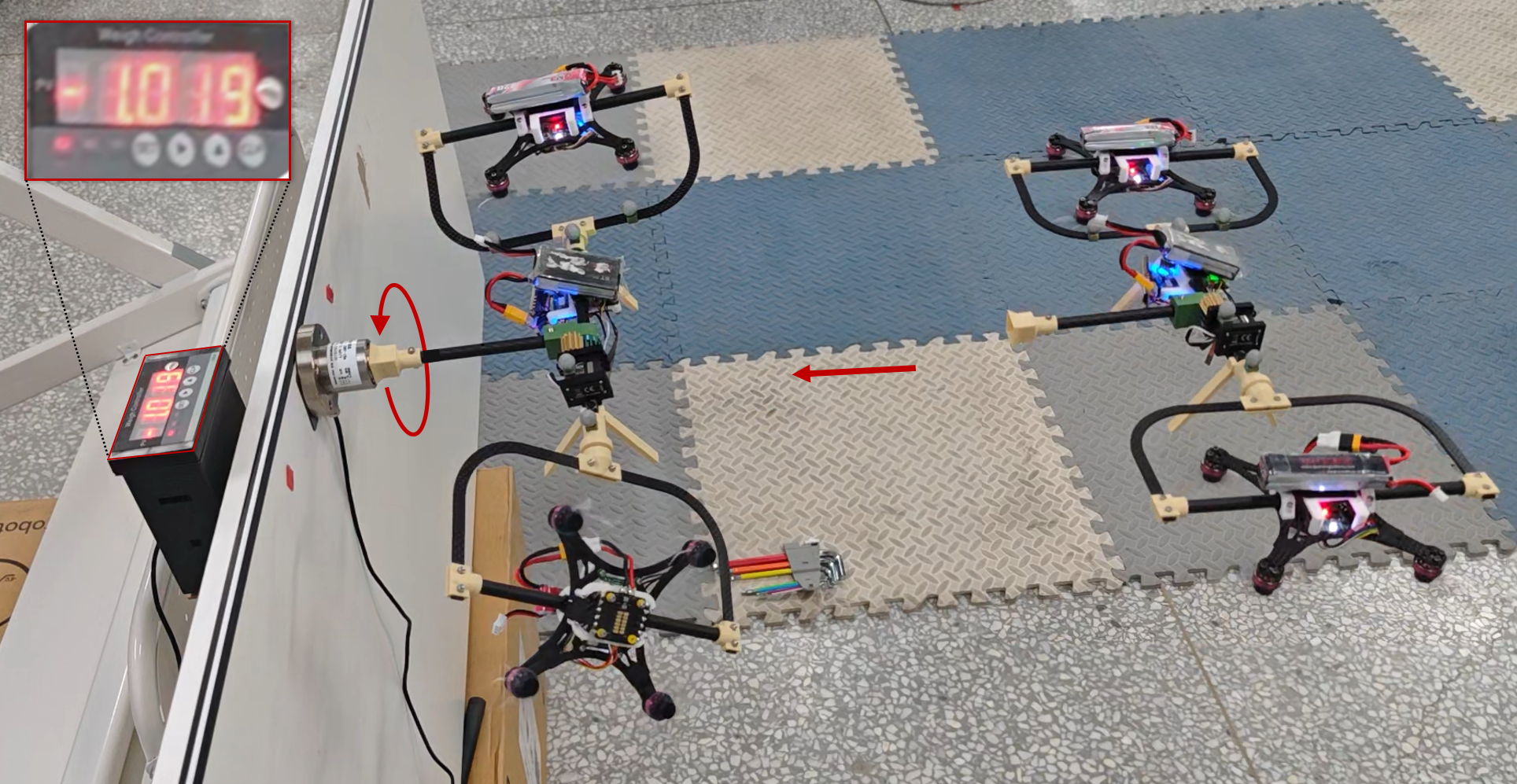}
\caption{Static torque generation test of FAM-DQ. While docked to a high-precision sensor, the system applies an increasing torque. The enlarged inset shows the maximum recorded absolute torque value of $1.019~\text{N}\cdot\text{m}$.}
\label{fig:exp3}
\end{figure}

To quantitatively evaluate the torque output capability of FAM-DQ, a test platform is established, as illustrated in Fig. \ref{fig:exp3}. As shown in the figure, FAM-DQ approaches the test platform until its distal end, equipped with an interface, aligns and embeds with a torque sensor which has a resolution of 0.001 $N \cdot m$. FAM-DQ then quasi-statically generates a gradually increasing torque about the $\boldsymbol{x}_e$-axis. The maximum measured absolute torque is $1.019~\text{N}\cdot\text{m}$. Given the total system mass of $0.447~\text{kg}$, this corresponds to a torque-to-mass ratio of approximately $2.28~\text{N}\cdot\text{m/kg}$,  demonstrating the platform's high torque-generation capability relative to its mass.

\subsection{Screw Driving}

\begin{figure}[t]
\centering
\subfloat[]{\includegraphics[width=0.48\linewidth]{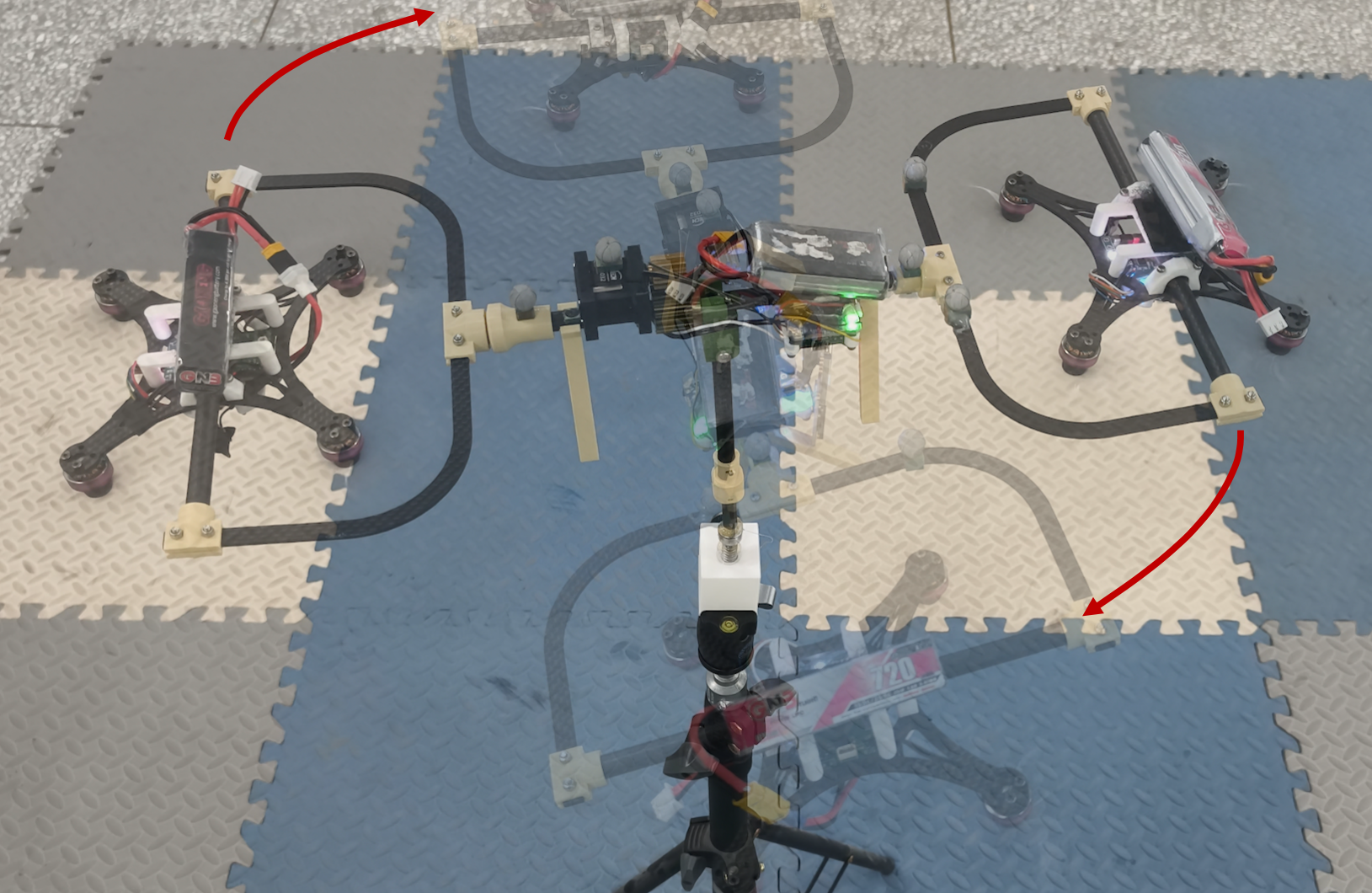}\label{fig:exp4_a}} \hfill
\subfloat[]{\includegraphics[width=0.48\linewidth]{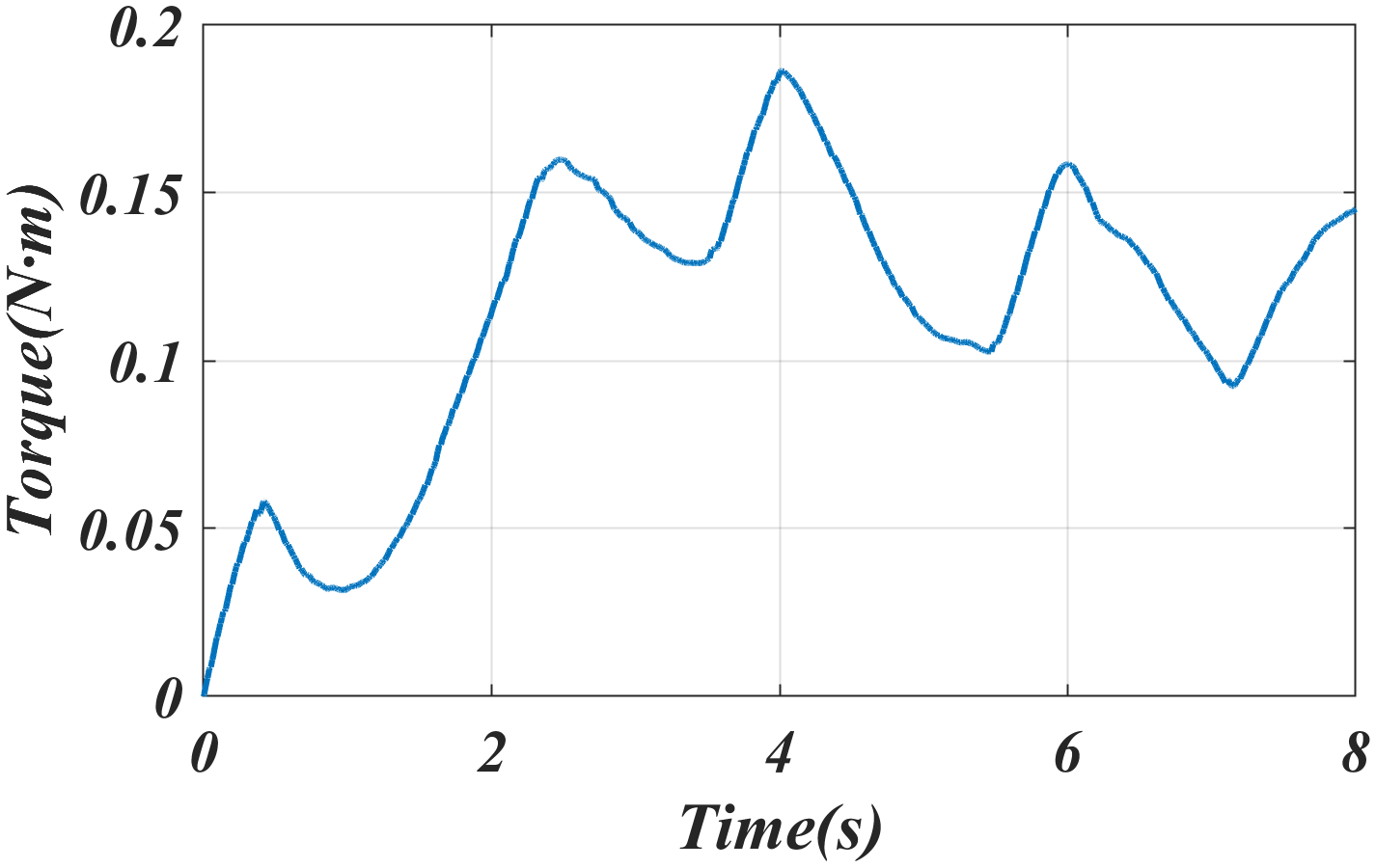}\label{fig:exp4_b}}
\caption{FAM-DQ drives a socket head cap screw into a rigid workpiece. (a) The overlaid transparent frames capture the continuous rotational motion, while the red arrows denote the direction of the torque applied along $\boldsymbol{x}_e$. (b) Torque generated by FAM-DQ during tightening.}
\label{fig:exp4}
\end{figure}

To evaluate the practical manipulation capability of FAM-DQ, a screw driving experiment is conducted, as shown in Fig. \ref{fig:exp4}. A screwdriver bit is mounted on the end-effector, and a M$8 \times 24$ mm socket head cap screw is temporarily attached to the bit. The target workpiece is rigidly fixed beneath the operating workspace.

During the experiment, FAM-DQ first approaches the target and aligns the screw with the hole. After contact, the platform applies a downward force while generating a torque about the $\boldsymbol{x}_e$-axis, as indicated in Fig. \ref{fig:exp4}\subref{fig:exp4_a}. The screw is successfully driven into the workpiece, and the estimated torque in Fig. \ref{fig:exp4}\subref{fig:exp4_b} confirms continuous torque generation during tightening. This result demonstrates the feasibility of FAM-DQ for contact-rich aerial manipulation tasks requiring both axial force and rotational torque.

\section{Conclusion}

This paper presented FAM-DQ, a dual-quadrotor fully actuated aerial manipulation platform designed for high-torque aerial physical interaction. By mounting two propulsion modules at the ends of a central frame through passive joints, FAM-DQ exploits the frame as a physical lever to amplify thrust-generated torque while maintaining independent control of the end-effector pose. The system dynamics were derived, and a lightweight 6-DoF controller was developed for decoupled position and attitude regulation.

The proposed platform was validated through circular trajectory tracking, hovering attitude tracking, static torque measurement, and screw-driving experiments. The results demonstrate that FAM-DQ can track spatial motion while maintaining a nearly constant attitude and regulate attitude while hovering at a fixed position. The screw-driving experiment further demonstrates the feasibility of using FAM-DQ for contact-rich aerial manipulation tasks.

Future work will focus on optimizing the mechanical design to improve torque output and energy efficiency, and on developing advanced control strategies for more complex manipulation tasks in dynamic environments.

\bibliographystyle{IEEEtran}
\bibliography{reference.bib}

\end{document}